%% file: main.tex
\documentclass{article}

\usepackage[dblblindworkshop,final]{neurips_2026}
\workshoptitle{SLMs for Agentic Systems, Paris, France, 2026 (extended version)}

\usepackage[utf8]{inputenc}
\usepackage[T1]{fontenc}
\usepackage[draft]{hyperref}
\usepackage{url}
\usepackage{booktabs}
\usepackage{amsfonts}
\usepackage{amsmath}
\usepackage{amssymb}
\usepackage{nicefrac}
\usepackage{microtype}
\usepackage{xcolor}
\usepackage{graphicx}
\usepackage{subcaption}
\usepackage{multirow}
\usepackage{wrapfig}

\title{E8 Lattice Key Quantization for Low-Bit KV-Cache Compression}

\author{%
  Sasho Nedelkoski \\
  Technical University of Berlin \\
  Berlin, Germany \\
  \texttt{sasho.nedelkoski@campus.tu-berlin.de} \\
  \And
  Alexander Acker \\
  logsight.ai \\
  Berlin, Germany \\
  \texttt{alexander.acker@logsight.ai} \\
  \AND
  Soeren Becker \\
  logsight.ai \\
  Berlin, Germany \\
  \texttt{soeren.becker@logsight.ai} \\
  \And
  Dominik Scheinert \\
  logsight.ai \\
  Berlin, Germany \\
  \texttt{dominik.scheinert@logsight.ai} \\
  \AND
  Odej Kao \\
  Technical University of Berlin \\
  Berlin, Germany \\
  \texttt{odej.kao@tu-berlin.de} \\
}

\begin{document}

\maketitle

\begin{abstract}
Compressing the key-value (KV) cache is essential for serving long-context large language models (LLMs) under memory constraints.
While current scalar quantization methods achieve significant compression, they typically face a quality-fidelity trade-off, either suffering from performance instability at 2-bit precision or necessitating 3-bit keys to maintain model quality. In this work, we demonstrate that a uniform low-bit k2v2 configuration (2-bit keys and 2-bit values) can preserve near-baseline quality by using vector quantization (VQ) based on the $E_8$ lattice for KV keys.
We show that structured VQ exploits correlations across adjacent dimensions to achieve signal-to-quantization-noise ratios (SQNR) higher than those of independent scalar quantization at the same key payload rate.
To ensure practical utility, we adapt a table-based asymmetric scoring algorithm that avoids explicit codebook dequantization during attention computation and preserves decode-throughput parity with scalar baselines. Across five LLM families, our $E_8$ k2v2 approach yields \textasciitilde$6\times$ KV compression, roughly 20\% smaller than scalar k3v2 (3-bit keys and 2-bit values), while retaining near-baseline long-context perplexity.
In our full-suite perplexity evaluation, $E_8$ improves over scalar k3v2 in 80\% of model--dataset--context settings.
These results indicate that $E_8$ lattice VQ improves quality per byte while keeping throughput at parity, suggesting that structured key quantization is a useful primitive for KV-compression stacks.
Overall, our work provides a low-overhead primitive for expanding the quality-memory frontier in long-context LLM inference.
\end{abstract}

\section{Introduction}
\label{sec:intro}
Supporting long context windows in large language models (LLMs) is essential for in-context learning, long-horizon reasoning, holistic codebase analysis, and retrieval-augmented generation~\citep{wang2024limitssurveytechniquesextend}.
LLM inference requires storing key-value (KV) pairs for every token in the context window across all attention layers and heads, placing high pressure on available device memory~\citep{li2025surveylargelanguagemodel}.
For a model with $L$ layers, $H_\text{kv}$ KV heads, and head dimension $d$, the KV cache consumes $2 \cdot L \cdot H_\text{kv} \cdot d \cdot 2$ bytes per token in fp16.
At 128K tokens of context, even an 8B grouped-query model requires tens of gigabytes of KV cache, and models with more KV heads can exceed the memory footprint of the model weights themselves.
This bottleneck limits batch size and context length, leading to prohibitive deployment cost.
Consequently, KV-cache efficiency directly affects whether long-context serving fits on a single accelerator or requires multi-GPU execution.

To mitigate these memory constraints, KV cache quantization has emerged as a prominent solution~\citep{li2025surveylargelanguagemodel}. 
Recent KV-cache quantization approaches, including KIVI~\citep{kivi}, KVQuant~\citep{kvquant}, and TurboQuant~\citep{turboquant}, rely primarily on per-channel scalar quantization. 
Treating each of the key dimensions $d$ as an independent signal, a quantization of $b$ bits assigns each dimension to one of the $2^b$ levels, achieving a rate of $b$ bits/dim.
However, scalar quantization often incurs quality degradation at very low bitrates.
Recent hardware-optimized configurations typically necessitate 3 bits for keys (e.g., TurboQuant's k3v2 setting: 3-bit keys and 2-bit values) to preserve model quality. 

We hypothesize that the degradation of low-bit scalar key quantization is partly due to limited use of geometric structure in key vectors.
While scalar methods are computationally efficient, they treat each dimension in isolation, ignoring the inter-dimensional correlations prevalent in LLM hidden states~\citep{engels2025languagemodelfeaturesonedimensionally}. 
To capture this structure, we use lattice VQ over blocks of dimensions to exploit space-filling and shaping gains, which can provide higher SQNR than scalar quantization at matched bit rates~\citep{zamir_lattice}.

Specifically, we employ the $E_8$ lattice in $\mathbb{R}^8$, which offers a useful combination of strong lattice geometry and practical efficiency. 
It achieves the densest possible sphere packing in eight dimensions~\citep{viazovska2017}, with near-spherical Voronoi cells that provide a shaping advantage over cubic scalar quantizers.
Our implementation uses a structured $E_8$-derived (E8P12-style) indexing scheme with parity/sign constraints and a two-branch $\pm 0.25$ selection rule, enabling efficient nearest-codeword assignment without brute-force search over all codewords.

\textbf{Contributions.} This paper makes the following contributions:
\begin{enumerate}
    \item We apply $E_8$ lattice VQ to key compression in the KV cache, partitioning each $d$-dimensional key vector into independent 8-dimensional blocks. At 2 bits/dim payload (65,536 codewords per block; 2.25 bits/dim effective key rate including norm metadata), the resulting k2v2 configuration achieves \textasciitilde$6\times$ total KV compression relative to fp16. Reported memory numbers are end-to-end KV-cache bytes/token, counting keys, values, quantized indices/signs, norms, and value scale/zero metadata.
    \item We show that $E_8$ at low key rate achieves rate-distortion performance between scalar 3-bit and 4-bit quantization. On 128-dimensional Gaussian vectors, $E_8$ obtains 10.47\,dB SQNR, compared with 7.37\,dB for scalar 3-bit, while using a lower effective key rate. In a controlled attention-fidelity probe, the 2-bit $E_8$ configuration also yields lower average attention KL divergence than scalar k3v2 (0.045 vs. 0.091) and KIVI k2v2 (0.224), indicating better preservation of fp16 attention distributions under compression.
    \item To avoid dequantizing vector-quantized keys during decoding, we adapt table-based asymmetric scoring to the fixed $E_8$ key codebook. For each query-head block, we precompute the dot product between the query block and all $65{,}536$ $E_8$ codewords. Each compressed key block can then be scored by indexing this table with its stored codeword id, replacing an 8-dimensional dot product with a lookup.
    This removes explicit key dequantization from the decode-time scoring path in our serving path and matches the throughput of optimized scalar decoders in the measured setting.
    \item We provide full-suite evaluations across five LLMs (Qwen2.5, Llama-3.1, Mistral, Falcon3, and Granite). 
    The results show that $E_8$ k2v2 improves the quality--memory trade-off over scalar k3v2 in most long-context perplexity settings and has lower perplexity than KIVI k2v2 across the evaluated perplexity suite, without a measurable decode-throughput penalty.
\end{enumerate}

\section{Related work}
\label{sec:related}


\paragraph{KV-cache quantization.}
Current KV-cache compression methods primarily use scalar quantization. KIVI~\citep{kivi} uses per-channel asymmetric key quantization and per-token value quantization; KVQuant~\citep{kvquant} adds pre-RoPE keys, non-uniform calibrated grids, and sparse outlier handling; GEAR~\citep{gear} combines low-rank approximation with quantized residuals; TurboQuant~\citep{turboquant} uses QJL random projections for calibration-free robustness, and its k3v2 setting is our primary scalar baseline. ZipCache~\citep{zipcache}, LeanKV~\citep{leankv}, and AQUA-KV~\citep{aquakv} further adapt bit-widths across tokens, keys/values, or predictable residual components. These methods are effective, but the dominant coding operation remains scalar or projection-based; our work instead encodes each key block with a structured geometry-aware codebook.

\paragraph{Joint coding and vector quantization.}
Coupled Quantization (CQ)~\citep{cq} observes that joint KV-channel entropy can be far below the sum of marginal entropies, motivating cross-channel coding; PM-KVQ~\citep{pmkvq} highlights the need for long-context-aware calibration under RoPE. Recent VQ-based KV-cache systems, including NestQuant~\citep{nestquant} and CommVQ~\citep{commvq}, further show that lattice or learned vector codebooks can improve low-bit cache compression. Our method shares the premise that key channels should not be treated independently, but differs by using a fixed, non-learned $E_8$ key codebook with online encoding and by adding decode-time table scoring so attention can operate directly on compressed keys. Related VQ work for weights, including PQ~\citep{jegou2011pq,guo2020opq}, QuIP~\citep{quip}, QuIP\#~\citep{quipsharp}, and AQLM~\citep{aqlm}, shows that structured codebooks can achieve strong low-bit rate-distortion. KV caches impose different constraints: activations are produced online during decoding and attention scores must be computed against a growing cache, so avoiding dense key reconstruction is central.

\paragraph{Token reduction and attention approximation.}
Token-eviction systems reduce cache length rather than bits per token. StreamingLLM~\citep{streamingllm} retains recent tokens and attention sinks; H2O~\citep{h20} and SnapKV~\citep{snapkv} retain tokens selected by accumulated or pooled attention; MagicPIG~\citep{magicpig} uses LSH and CPU-resident hash tables for sparse long-context attention. Projection methods such as Loki~\citep{loki} reduce effective key dimensionality before attention. These approaches are orthogonal to our quantizer: one could apply $E_8$ key coding to the retained token set or to a projected subspace.

\paragraph{Lattice codes.}
Lattice quantization theory~\citep{zamir_lattice,conway_sloane} shows that structured lattices achieve space-filling gain that scalar quantizers cannot. The $E_8$ lattice is the densest known sphere packing in eight dimensions~\citep{viazovska2017}; its 240 kissing neighbors and near-spherical Voronoi cells give a shaping advantage over cubic lattices and motivate its use in low-bit neural compression. To our knowledge, this is the first application of $E_8$ coding to runtime KV caches with online encoding and throughput-neutral compressed-domain attention scoring.

\section{Method}
\label{sec:method}

\subsection{Preliminary baseline}
As a point of comparison, standard scalar KV-cache quantization independently quantizes each key dimension.
Consider a key vector $\mathbf{k} \in \mathbb{R}^d$, where $d$ is the attention head dimension.
Scalar quantization at $b$ bits independently quantizes each dimension:
\begin{equation}
    \hat{k}_i = s \cdot \text{round}\!\left(\frac{k_i}{s}\right), \quad s = \frac{\max_i |k_i|}{2^{b-1} - 1},
    \label{eq:scalar}
\end{equation}
storing $b \cdot d$ key-index bits plus scale metadata.
Eq.~\eqref{eq:scalar} is a generic per-vector scalar-quantization reference used to introduce notation; concrete KV baselines may use per-channel, per-token, or groupwise scaling. Our implemented scalar baseline (TurboQuant k3v2) uses rotated unit-norm keys with a Lloyd-Max MSE stage plus a 1-bit QJL residual-sign stage.
Throughout the paper, we assume that $d$ is divisible by 8, and as our running example we choose $d=128$ to make the explanation clear. However, the framework generalizes to arbitrary head dimensions satisfying this constraint. Compression ratios refer to \emph{effective KV-cache bytes per token}, not nominal key bitwidth alone.
A complete fp16 KV cache stores one key and one value vector per token-head, requiring $2 \cdot d \cdot 16$ bits. As an example, for $d=128$, this is $4096$ bits per token-head.
As an example, the TurboQuant k3v2~\citep{turboquant} setting stores $3d$ packed key bits and $2d$ packed value bits, plus metadata for dequantization and compressed-domain scoring:
\[
    B_{\mathrm{S3}} = 3d + 2d + m_{\mathrm{S3}}.
\]
End-to-end accounting includes packed keys, packed values, and per-token metadata such as QJL sign bits, norms, and value scale/zero parameters; it excludes fixed model-level state such as rotation matrices and shared codebooks.
According to this convention, the k3v2 scalar yields about ${\sim}4.8\times$ compression over fp16, as shown in Table~\ref{tab:memory} below.
\subsection{E8 lattice vector quantization for keys}
The $E_8$ lattice $\Lambda_{E_8} \subset \mathbb{R}^8$ is defined as
\begin{equation}
    \Lambda_{E_8} = \left\{ \mathbf{x} \in \mathbb{Z}^8 \cup (\mathbb{Z} + \tfrac{1}{2})^8 \;\Big|\; \sum_{i=1}^8 x_i \equiv 0 \pmod{2} \right\}.
\end{equation}
The full $E_8$ lattice underlies the optimal sphere packing in 8D space~\citep{viazovska2017}, making it a natural high-dimensional quantization template.
We use a fixed, non-learned $E_8$-derived codebook $\mathcal{C}\subset\mathbb{R}^8$ with $2^{16}=65{,}536$ codewords, so each 8-dimensional key block is represented by one 16-bit index. The codebook is built from an $E_8$-consistent half-integer grid augmented with norm-12 patterns. Its indexing structure separates an 8-bit absolute-value prototype from an 8-bit sign/parity pattern, enabling deterministic encode/decode without brute-force search over all codewords. During encoding, we evaluate two $\pm0.25$ shifted lattice branches and store the lower-error index.
This work focuses on the fixed $E_8$ key codec itself and does not add a separate sparse-outlier retention path. Such outlier handling is complementary and left for future work.
\paragraph{Encoding.}
To quantize a key vector $\mathbf{k} \in \mathbb{R}^{d}$, the method applies a random orthogonal rotation matrix $\mathbf{R} \in \mathbb{R}^{d \times d}$ (shared across all keys in a layer and fixed at initialization) to spread energy across coordinates and mitigate channel-aligned outliers while preserving $L_2$ distances.
Following the assumption that $d$ is a multiple of $8$ (which is the case in most modern LLMs ~\citep{wang2025parametersperformancedatadrivenstudy}), we then partition the rotated vector into $M=\frac{d}{8}$ discrete 8-dimensional blocks.
For $d=128$, this gives $M=16$ blocks:
\begin{equation}
    \tilde{\mathbf{k}} = \mathbf{R}\mathbf{k}, \quad \tilde{\mathbf{k}} = [\tilde{\mathbf{k}}^{(1)}, \ldots, \tilde{\mathbf{k}}^{(16)}], \quad \tilde{\mathbf{k}}^{(j)} \in \mathbb{R}^8.
\end{equation}
Each block is scaled and quantized to its nearest $E_8$ codeword:
\begin{equation}
    z_j = \arg\min_{i \in \{0,\ldots,|\mathcal{C}|-1\}} \left\| \frac{\alpha}{\|\tilde{\mathbf{k}}\|_2} \tilde{\mathbf{k}}^{(j)} - \mathbf{g}_i \right\|_2^2, \quad j = 1, \ldots, 16,
    \label{eq:codeword}
\end{equation}
where $\alpha$ is a calibrated scale parameter. In all main experiments, $\alpha$ is selected deterministically by minimizing block reconstruction MSE on 300 seeded Gaussian unit vectors using a coarse scale grid followed by local refinement; for $d=128$ this gives $\alpha\approx12$.
The compressed key is stored as $16 \times 16$ index bits plus one 32-bit norm $\|\tilde{\mathbf{k}}\|_2$, totaling $16 \times 16 + 32 = 288$ bits $= 36$ bytes, yielding a rate of $288/128 = 2.25$ bits/dim.
\paragraph{Effective rate.}
For $d=128$, the raw packed $E_8$ k2v2 payload is $288$ key bits plus $2d=256$ value bits, or $544$ bits per token-head before value metadata.
Under the accounting convention above, this gives the ${\sim}6\times$ end-to-end KV-cache compression over fp16 (Table~\ref{tab:memory}).
\subsection{Table-based asymmetric scoring}
\label{sec:table_scoring}
A naive implementation of attention with lattice VQ keys would dequantize each key by looking up its $M$ codewords from the $E_8$ grid, reconstructing a dense $d$-dimensional vector, then computing the dot product with the query.
For $N$ cached keys, this requires $NM$ random accesses to a $65{,}536 \times 8$ float32 table---a memory bandwidth bottleneck that typically renders lattice VQ slower than scalar quantization.
We instead use \emph{table-based asymmetric scoring}, following asymmetric-distance computation in product quantization~\citep{jegou2011pq} and recent lookup-based KV-attention systems~\citep{pqcache,polarquant,lookat}.
Given a query $\mathbf{q} \in \mathbb{R}^d$ for a single decoding step:
\begin{enumerate}
    \item Rotate and partition: We apply the same orthogonal rotation to the query $\tilde{\mathbf{q}} = \mathbf{R}\mathbf{q}$, and partition it into $M$ blocks $\tilde{\mathbf{q}}^{(j)} \in \mathbb{R}^8$ (e.g., $M=16$ for $d=128$).
    \item Pre-compute lookups: We build $M$ lookup tables: $T_j[i] = \langle \tilde{\mathbf{q}}^{(j)}, \mathbf{g}_i \rangle$ for $i = 0, \ldots, 65{,}535$, where $\mathbf{g}_i$ is the $i$-th $E_8$ codeword. Each of the $M$ tables contains $65{,}536$ float32 entries.
    \item Asymmetric scoring: For each cached key stored as $M$ indices $(z_1, \ldots, z_{M})$ and norm $n=\|\tilde{\mathbf{k}}\|_2$, we compute the attention score using the calibrated parameter $\alpha$ from Eq.~\eqref{eq:codeword}:
    \begin{equation}
        \text{score} = \frac{n}{\alpha} \sum_{j=1}^{M} T_j[z_j].
    \end{equation}
\end{enumerate}
\paragraph{Computational efficiency.}
Step~2 costs $M \times 65{,}536$ 8-dimensional dot products, implemented as a batched matrix multiplication over query blocks and codewords.
Step~3 costs $NM$ table gathers plus $NM$ additions.
The $M$ lookup tables require $M \times 65,536 \times 4$ bytes of memory.
For $M=16$, this totals $4$\,MB per query head.
Because each gather is a single indexed read from a contiguous query-local table, per-key scoring is memory-efficient in our decode setting; only the once-per-query table build scales with codebook size.
For batched teacher-forcing quality evaluations (perplexity and downstream tasks), we use the same compressed representation and asymmetric inner-product estimator. These batched evaluations use a mathematically equivalent dense reconstruction when processing many query positions at once; the decode throughput results in Table~\ref{tab:throughput} exercise the table-based serving path directly.
\subsection{Implementation details}
We integrate $E_8$ key quantization into the TurboQuant~\citep{turboquant} framework, retaining its value quantization and residual window for recently inserted tokens. The reported $E_8$ k2 key path does not use QJL residual signs; QJL is retained only in scalar TurboQuant baselines.
The random orthogonal rotation matrix $\mathbf{R}\in\mathbb{R}^{d \times d}$ is initialized once per layer from a seeded PRNG.
For $d=128$, storing this dense matrix in fp32 requires 64\,KB per layer (or 32\,KB in fp16), which is a negligible model-level overhead in our accounting.
Encoding a single key requires one matrix-vector multiply followed by $M$ fast structured nearest-codeword searches over 8D blocks. 
Rather than brute-force $O(|\mathcal{C}|)$ VQ lookup, the implementation first rounds in a reduced absolute-value candidate set, then applies sign/parity-consistent index reconstruction with the $\pm0.25$ branch selection.
As shown in the rate-distortion analysis in Figure~\ref{fig:rate_distortion}, this integration yields an $E_8$ configuration whose SQNR lies between scalar 3-bit and 4-bit baselines while remaining in the low-bit key-rate regime.

\section{Experiments}
\label{sec:experiments}

\subsection{Setup}

\paragraph{Models.}
We evaluate on five instruction-tuned LLMs spanning different architectures and labs: Qwen2.5-14B~\citep{qwen2025qwen25technicalreport}, Mistral-7B~\citep{jiang2023mistral7b}, Llama-3.1-8B~\citep{grattafiori2024llama3herdmodels}, Falcon3-7B~\citep{almazrouei2023falconseriesopenlanguage}, and Granite-3.1-8B~\citep{stallone2024scalinggranitecodemodels}.
All models use grouped-query attention with head dimension divisible by 8 and varying numbers of KV heads.

\paragraph{Configurations.}
We compare four KV cache configurations: (1) Baseline: fp16 keys and values; (2) KIVI k2v2~\citep{kivi}; (3) Scalar k3v2 (TurboQuant~\citep{turboquant}); and (4) our E8 Lattice k2v2 setting.
All compressed configurations use 2-bit values and a recent-token residual buffer of 128 tokens unless explicitly varied in residual-window sweeps.

\begin{wrapfigure}{r}{0.7\textwidth}
    \vspace{-12pt}
    \centering
    \includegraphics[width=0.69\textwidth]{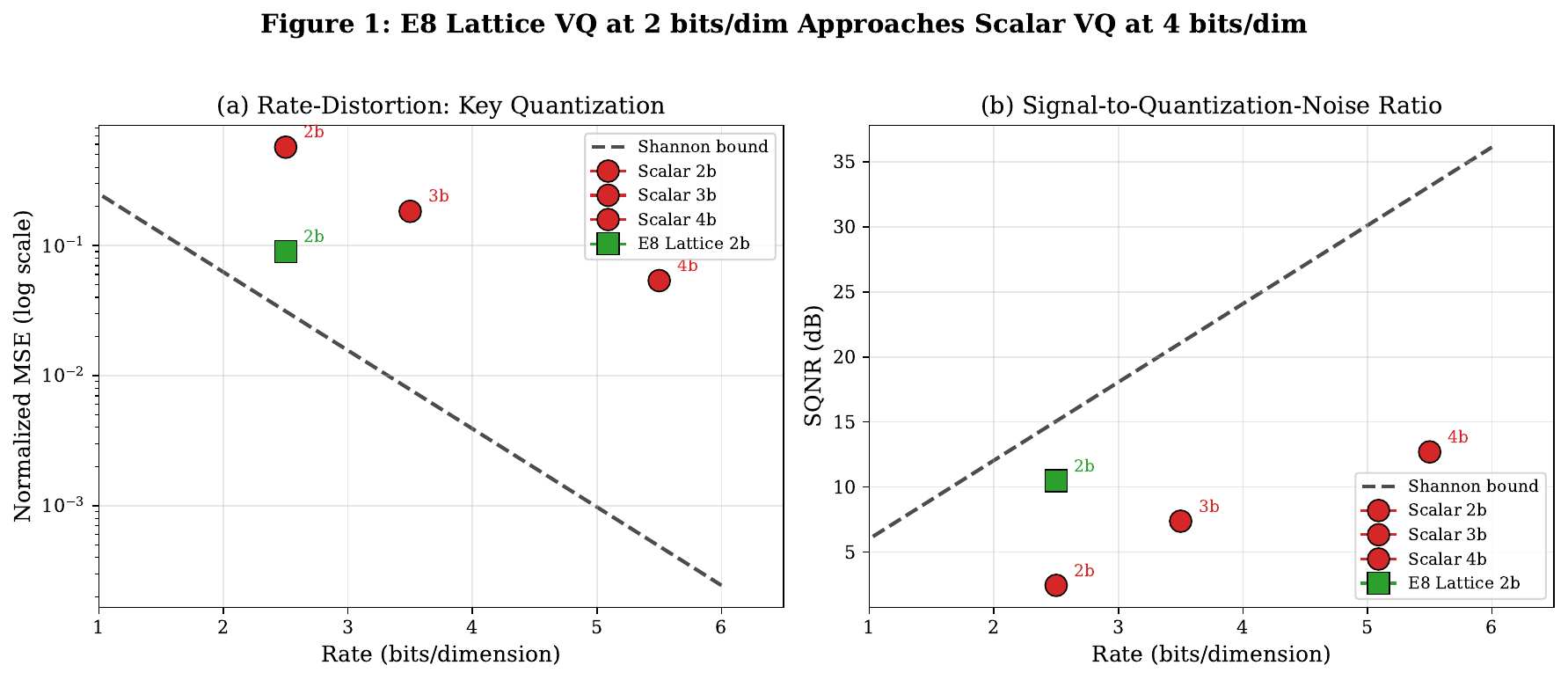}
    \caption{Rate-distortion comparison on 128-dimensional Gaussian vectors. $E_8$ lattice VQ achieves SQNR between scalar 3-bit and 4-bit quantization at a comparable low key rate.}
    \label{fig:rate_distortion}
    \vspace{-8pt}
\end{wrapfigure}

\paragraph{Metrics.}
To provide a holistic assessment, we categorize our evaluation into three axes:
(1) End-to-end quality: long-context perplexity on WikiText-2~\citep{merity2016pointersentinelmixturemodels} and C4~\citep{dodge2021documentinglargewebtextcorpora} (contexts 512, 2048, 4096), downstream few-shot accuracy on HellaSwag (10-shot)~\citep{zellers2019hellaswagmachinereallyfinish}, ARC-Challenge (25-shot)~\citep{clark2018thinksolvedquestionanswering}, and WinoGrande (5-shot)~\citep{sakaguchi2019winograndeadversarialwinogradschema}, alongside a needle-in-a-haystack (NIAH) retrieval sanity check.
(2) Internal fidelity: per-layer cosine similarity between compressed and fp16 key representations, and KL divergence for attention distributions.
(3) System performance: single-sequence decode throughput (tokens/s) on a single 80 GB NVIDIA A100 GPU and total memory consumption.
We report single-system, single-sequence serving measurements; larger-batch scheduling behavior, cross-hardware variance, and cross-backend variance are left to future work.

\subsection{Rate-distortion analysis}

Figure~\ref{fig:rate_distortion} shows rate-distortion curves for our implemented TurboQuant-style scalar/product key baseline and the $E_8$ lattice key quantizer on 128-dimensional Gaussian vectors.
At the low-rate operating point, $E_8$ achieves SQNR of 10.47\,dB versus 2.43\,dB for the 2-bit scalar/product baseline.
More practically, $E_8$ at 2 bits/dim (SQNR 10.47\,dB, cosine similarity 0.954) falls \emph{between} the implemented scalar 3-bit (7.37\,dB, 0.920) and 4-bit (12.69\,dB, 0.974) configurations.
This comparison supports the empirical quality-per-bit advantage of the fixed $E_8$ key codec in our compression stack; it should not be read as an optimal scalar-quantizer lower bound.

\subsection{Perplexity}

\input{table1_perplexity}

Table~\ref{tab:perplexity} shows complete WikiText-2 and C4 perplexity at contexts 512/2048/4096 for all five full-suite models.
$E_8$ k2v2 improves over scalar k3v2 in 24/30 model--dataset--context blocks while using fewer key bits; these wins include every non-Falcon model and both corpora.
The gains are most visible at longer context lengths, where quantization error accumulates over more cached keys.
For example, at 4K context on C4, Qwen2.5-14B degrades by only +0.29 perplexity under $E_8$, compared with +1.41 for scalar k3v2 and +5.20 for KIVI; Llama-3.1-8B shows the same pattern (+0.52 for $E_8$, +1.17 for scalar, +36.54 for KIVI).
Granite-3.1-8B also benefits substantially at 4K, with WikiText-2/C4 deltas of +0.30/+0.50 for $E_8$ versus +0.69/+0.94 for scalar.

The only systematic scalar advantage appears on Falcon3-7B, where scalar k3v2 is best in all six Falcon rows.
Even there, the gap is small in absolute perplexity (at most 0.22), and $E_8$ remains below KIVI in every Falcon row.
This suggests that Falcon's activation statistics are unusually favorable to per-coordinate scalar quantization, but not that the 2-bit lattice representation is generally brittle.
Overall, the robust conclusion is that $E_8$ moves the quality--memory Pareto frontier beyond scalar k3v2 for most evaluated long-context settings and dominates KIVI k2v2 across this perplexity suite.

\subsection{Memory compression}

\input{table2_memory}
Table~\ref{tab:memory} quantifies per-token KV cache memory.
$E_8$ k2v2 achieves ${\sim}6\times$ effective compression for the four $H_{\text{kv}}=8$ models, compared with ${\sim}4.8\times$ for scalar k3v2.
This corresponds to roughly a 20\% smaller compressed cache than scalar k3v2 after accounting for packed key/value payloads and per-token metadata.
The Falcon3-7B row is slightly different because its larger head dimension better amortizes fixed per-vector metadata such as key norms, yielding 5.1$\times$ for scalar and 6.2$\times$ for $E_8$.

These ratios use the effective bytes/token accounting defined in Section~\ref{sec:method}.
For a representative 8B architecture such as Llama-3.1-8B, scalar k3v2 requires 27.3\,KB/token, whereas $E_8$ k2v2 requires 21.7\,KB/token, saving 5.6\,KB per token in the quantized cache region.
At 128K context length, this accounting projects roughly 700\,MB of additional memory saved per sequence.

\subsection{Decode throughput}

Table~\ref{tab:throughput} demonstrates that table-based asymmetric scoring (detailed in Section~\ref{sec:table_scoring}) successfully eliminates the computational bottleneck traditionally associated with lattice VQ.
Across all five evaluated architectures and context lengths, $E_8$ k2v2 achieves decode-throughput parity with both the uncompressed fp16 baseline and the optimized scalar (k3v2) and KIVI kernels in our single-sequence decode benchmark.
The variance across all methods falls within the margin of standard measurement noise (±0.2\,tok/s), indicating no measurable latency degradation in this setting.
This empirical parity supports our system design in the measured setting: precomputing the lookup table avoids dense key reconstruction during attention, and the remaining table-build/gather overhead is below measurement noise for our single-sequence A100 decode path.
These throughput results are measured on our A100 serving path; additional validation on other accelerators and paged-attention runtimes is left for future work.

\input{table3_throughput}
\subsection{Downstream task accuracy}

\input{table4_downstream}

To ensure that extreme KV cache compression does not degrade foundational model capabilities, we evaluate zero-shot and few-shot reasoning performance on HellaSwag, ARC-Challenge, and WinoGrande. 
Table~\ref{tab:downstream} reports these downstream accuracies. 
Across all models, $E_8$ k2v2 strictly maintains parity with both the fp16 baseline and the scalar baselines.
In few-shot settings, much of the demonstration context is still quantized; the observed parity therefore suggests that the compressed cache preserves the in-context signal needed by these tasks, while the recent-token residual window protects the final local context.

\begin{figure}[t]
    \centering
    \includegraphics[width=0.55\columnwidth]{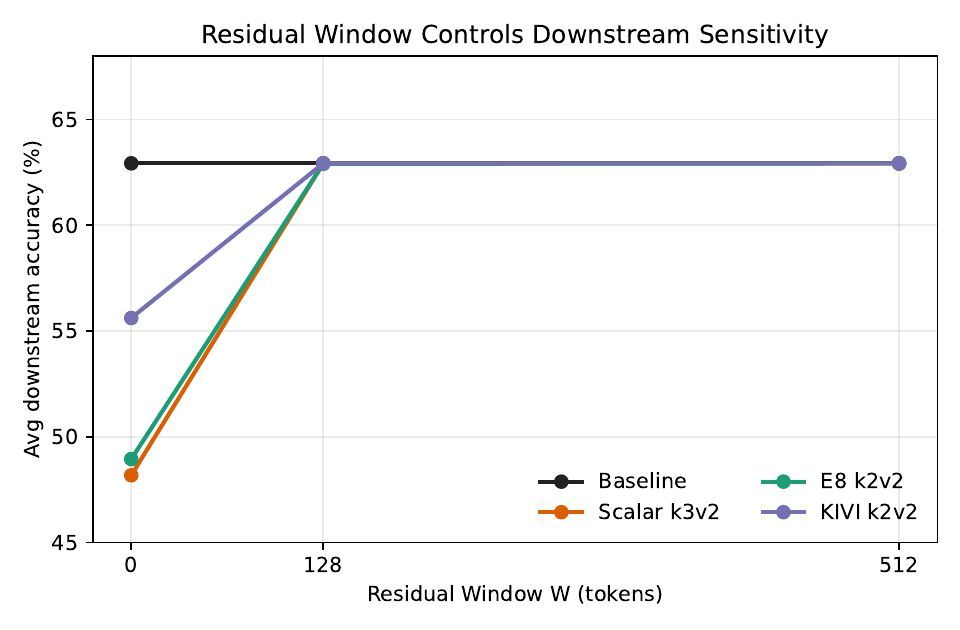}
    \caption{Average downstream accuracy across the core residual-sweep models versus residual window size. At $W=0$, compression quality differences are visible; by $W=128$, methods are nearly indistinguishable.}
    \label{fig:residual_window_sweep}
\end{figure}

To isolate the role of this residual window, we vary its size $W$ (Figure~\ref{fig:residual_window_sweep}). 
When the uncompressed window is entirely removed (W=0), the underlying full quantization noise is exposed, and distinct performance ordering reappears.
As $W$ expands to 128 tokens, the performance of all compressed configurations rapidly converges to the fp16 baseline, remaining practically indistinguishable for larger windows. 

\subsection{Attention fidelity}
\label{sec:attention_fidelity}

\begin{figure}[t]
    \centering
    \includegraphics[width=0.96\textwidth]{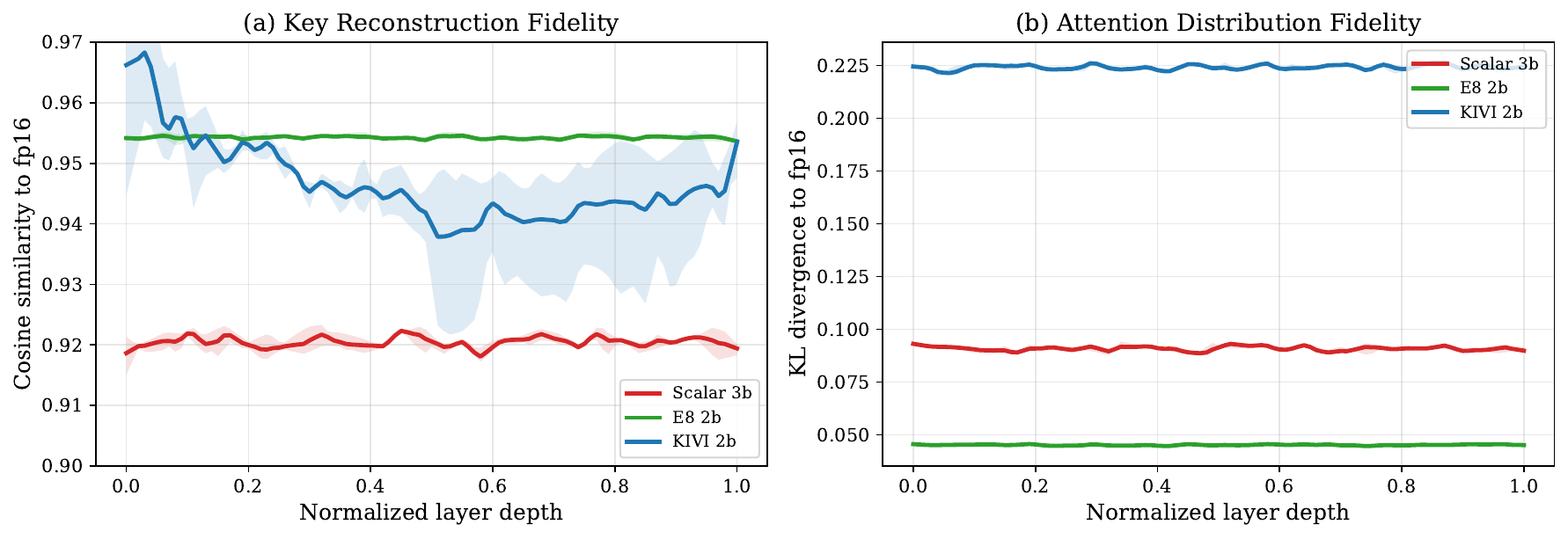}
    \caption{Left: per-layer cosine similarity between compressed and fp16 keys captured from model forward passes. Right: per-layer KL divergence in a controlled attention-distribution probe. Lines show mean over models and shaded bands show the min--max range after normalizing each architecture's layer index to $[0,1]$. }
    \label{fig:fidelity_summary}
\end{figure}

Figure~\ref{fig:fidelity_summary} shows per-layer cosine similarity between compressed and fp16 key representations.
$E_8$ at 2 bits/dim remains substantially higher than scalar at 3 bits (roughly 0.954 vs. 0.921 average), consistent with the rate-distortion analysis.
The shaded range is narrow and smooth across depth, suggesting lattice quantization error is not concentrated in a small subset of layers.

This reconstruction fidelity directly translates to superior preservation of the attention mechanism itself. 
Figure~\ref{fig:fidelity_summary}b illustrates the per-layer KL divergence between attention distributions computed with compressed versus fp16 keys. 
Here, $E_8$ achieves a remarkably stable and low KL divergence (0.045), effectively halving the error of the 3-bit scalar baseline (0.091). 
Furthermore, $E_8$ outperforms KIVI k2v2, which suffers from attention distribution shifts (KL > 0.200). 
This confirms that our lattice geometry preserves the true pre-softmax attention logits more faithfully than existing 2-bit methods, ensuring that improved geometric reconstruction meaningfully carries through to downstream attention behavior.

\subsection{Long-context retrieval robustness}

To verify that extreme KV cache quantization does not degrade the model's ability to fetch isolated facts from extended contexts, we conducted a standard single-needle-in-a-haystack retrieval evaluation similar to~\citep{turboquant}. 
We evaluated retrieval accuracy using a randomly placed fact injected into distractor contexts of varying lengths (4K, 8K, and 16K tokens) across all five models.

$E_8$ k2v2 achieved 100\% retrieval accuracy across all lengths and models, strictly matching the performance of the uncompressed fp16 baseline as well as the competing KIVI and scalar compression methods.

\subsection{Ablation studies}
\input{table5_ablation}

\paragraph{(a) $E_8$ scale calibration.}
The scale parameter $\alpha$ controls the dynamic range mapped to the $E_8$ quantizer. 
While values that are too small ($\alpha \leq 4$) cause strong clipping and values too large ($\alpha \geq 20$) waste codebook resolution, performance is remarkably robust within the intermediate range. 
We observe a wide, stable plateau between $\alpha=8$ and $\alpha=15$ where MSE and cosine similarity remain stable. 

\paragraph{(b) Mixed precision.}
We also test simple mixed-precision variants that change the first few layers. 
These variants do not improve the average cosine metric over uniform $E_8$ k2, so our main configuration keeps a uniform key format across layers.
Because the E8 3-bit variant adds a QJL residual correction rather than serving as a pure higher-fidelity E8 codebook, we interpret this ablation only as an implementation-level check, not as evidence that early layers never benefit from additional capacity.

\section{Conclusion}
\label{sec:conclusion}

We demonstrate that $E_8$ lattice vector quantization is a practical primitive for low-bit KV-cache key compression. Its fixed geometry-aware codebook enables table-based asymmetric scoring: for each query, dot products with the shared codebook are precomputed once and cached keys are scored directly from indices, avoiding explicit key dequantization in the decode path. Across diverse LLM families and long-context evaluations, this design improves key fidelity and quality per byte while preserving downstream quality and single-sequence decode-throughput parity. At the same time, making compressed-domain scoring efficient beyond this setting, especially for larger batches and prefill-heavy workloads remains an important systems challenge. These results suggest that structured vector codes are a useful building block for KV-compression systems.

{
\small
\bibliographystyle{plainnat}
\bibliography{bibliography}
}

\end{document}

%% file: table1_perplexity.tex
\begin{table}[t]
\centering
\caption{Perplexity on WikiText-2 and C4 across context lengths. Each block reports compressed-method perplexity deltas against fp16 baseline; lower is better. Bold marks the best compressed configuration within each dataset/model/context.}
\label{tab:perplexity}
\resizebox{\columnwidth}{!}{%
\begin{tabular}{l c c c c c c c c c c c c c c c c}
\toprule
\textbf{Model} & \textbf{Ctx} & \multicolumn{7}{c}{\textbf{WikiText-2}} & \multicolumn{7}{c}{\textbf{C4}} \\
\cmidrule(lr){3-9} \cmidrule(lr){10-16}
 & & \textbf{Base} & \textbf{KIVI} & \textbf{$\Delta$} & \textbf{Scalar} & \textbf{$\Delta$} & \textbf{E8} & \textbf{$\Delta$} & \textbf{Base} & \textbf{KIVI} & \textbf{$\Delta$} & \textbf{Scalar} & \textbf{$\Delta$} & \textbf{E8} & \textbf{$\Delta$} \\
\midrule
Qwen2.5-14B & 512 & 5.90 & 7.16 & +1.26 & 6.29 & +0.39 & \textbf{6.13} & +0.23 & 11.11 & 12.47 & +1.36 & 11.59 & +0.48 & \textbf{11.35} & +0.24 \\
 & 2K & 5.01 & 7.32 & +2.31 & 5.61 & +0.60 & \textbf{5.29} & +0.28 & 10.32 & 13.82 & +3.50 & 11.31 & +0.99 & \textbf{10.60} & +0.29 \\
 & 4K & 4.84 & 7.88 & +3.04 & 5.51 & +0.68 & \textbf{5.13} & +0.29 & 10.18 & 15.38 & +5.20 & 11.59 & +1.41 & \textbf{10.47} & +0.29 \\
\midrule
Llama-3.1-8B & 512 & 7.33 & 9.40 & +2.07 & 7.70 & +0.36 & \textbf{7.51} & +0.18 & 11.34 & 15.00 & +3.65 & 11.93 & +0.59 & \textbf{11.71} & +0.37 \\
 & 2K & 6.45 & 15.91 & +9.45 & 6.97 & +0.51 & \textbf{6.67} & +0.21 & 10.90 & 32.72 & +21.82 & 11.84 & +0.94 & \textbf{11.35} & +0.45 \\
 & 4K & 6.30 & 21.47 & +15.17 & 6.90 & +0.60 & \textbf{6.54} & +0.23 & 10.88 & 47.43 & +36.54 & 12.05 & +1.17 & \textbf{11.40} & +0.52 \\
\midrule
Mistral-7B & 512 & 5.75 & 6.25 & +0.50 & 5.85 & +0.10 & \textbf{5.81} & +0.07 & 8.87 & 9.51 & +0.64 & 8.99 & +0.12 & \textbf{8.93} & +0.06 \\
 & 2K & 4.98 & 5.93 & +0.94 & 5.12 & +0.14 & \textbf{5.06} & +0.07 & 8.21 & 9.72 & +1.51 & 8.42 & +0.21 & \textbf{8.28} & +0.07 \\
 & 4K & 4.84 & 6.12 & +1.27 & 5.00 & +0.16 & \textbf{4.92} & +0.08 & 8.10 & 10.25 & +2.15 & 8.35 & +0.25 & \textbf{8.18} & +0.08 \\
\midrule
Falcon3-7B & 512 & 8.91 & 9.40 & +0.49 & \textbf{8.87} & -0.04 & 9.08 & +0.17 & 17.85 & 18.53 & +0.68 & \textbf{17.89} & +0.04 & 17.96 & +0.11 \\
 & 2K & 7.19 & 7.79 & +0.60 & \textbf{7.08} & -0.11 & 7.30 & +0.11 & 15.94 & 17.12 & +1.19 & \textbf{15.82} & -0.11 & 15.98 & +0.04 \\
 & 4K & 6.63 & 7.16 & +0.53 & \textbf{6.57} & -0.06 & 6.72 & +0.09 & 15.16 & 16.67 & +1.51 & \textbf{15.19} & +0.03 & 15.24 & +0.08 \\
\midrule
Granite-3.1-8B & 512 & 5.53 & 6.68 & +1.15 & 5.90 & +0.36 & \textbf{5.69} & +0.16 & 8.49 & 9.85 & +1.36 & 8.85 & +0.36 & \textbf{8.68} & +0.19 \\
 & 2K & 4.97 & 7.22 & +2.25 & 5.52 & +0.55 & \textbf{5.19} & +0.22 & 8.25 & 11.57 & +3.33 & 8.98 & +0.74 & \textbf{8.59} & +0.35 \\
 & 4K & 4.89 & 8.11 & +3.22 & 5.58 & +0.69 & \textbf{5.19} & +0.30 & 8.34 & 13.14 & +4.79 & 9.28 & +0.94 & \textbf{8.85} & +0.50 \\
\bottomrule
\end{tabular}
}
\end{table}

%% file: table2_memory.tex
\begin{table}[t]
\centering
\caption{Effective compressed KV-cache bytes per token for the quantized cache region. The accounting includes packed key/value data and per-token metadata (key norms/QJL signs and value scale/zero groups), but excludes fixed model-level tables/rotations and the finite recent-token residual window.}
\label{tab:memory}
\begin{tabular}{l c c c c c c c}
\toprule
\textbf{Model} & \textbf{$L$} & \textbf{$H_{\text{kv}}$} & \textbf{Baseline} & \multicolumn{2}{c}{\textbf{Scalar k3v2}} & \multicolumn{2}{c}{\textbf{E8 Lattice k2v2}} \\
\cmidrule(lr){5-6} \cmidrule(lr){7-8}
 & & & (B/tok) & (B/tok) & Ratio & (B/tok) & Ratio \\
\midrule
Qwen2.5-14B & 48 & 8 & 196,608 & 40,944 & 4.8$\times$ & \textbf{32,592} & \textbf{6.0$\times$} \\
Llama-3.1-8B & 32 & 8 & 131,072 & 27,296 & 4.8$\times$ & \textbf{21,728} & \textbf{6.0$\times$} \\
Mistral-7B & 32 & 8 & 131,072 & 27,296 & 4.8$\times$ & \textbf{21,728} & \textbf{6.0$\times$} \\
Falcon3-7B & 28 & 4 & 114,688 & 22,652 & 5.1$\times$ & \textbf{18,452} & \textbf{6.2$\times$} \\
Granite-3.1-8B & 40 & 8 & 163,840 & 34,120 & 4.8$\times$ & \textbf{27,160} & \textbf{6.0$\times$} \\
\bottomrule
\end{tabular}
\end{table}

%% file: table3_throughput.tex
\begin{table}[htbp]
\centering
\setlength{\tabcolsep}{4.5pt}            
\renewcommand{\arraystretch}{0.88}       
\caption{Decode throughput (tok/s) with table-based asymmetric scoring. E8 Lattice matches Scalar throughput despite using structured VQ, thanks to lookup-table scoring that avoids codebook dequantization.}
\label{tab:throughput}
\begin{tabular}{l c c c c c}
\toprule
\textbf{Model} & \textbf{Context} & \textbf{Baseline} & \textbf{KIVI k2v2} & \textbf{Scalar k3v2} & \textbf{E8 Lattice k2v2} \\
\midrule
Qwen2.5-14B & 1K & 49.4 & 49.4 & 49.4 & 49.4 \\
 & 4K & 48.4 & 48.4 & 48.4 & 48.4 \\
\midrule
Llama-3.1-8B & 1K & 89.0 & 88.9 & 88.9 & 89.0 \\
 & 4K & 86.9 & 87.0 & 87.0 & 87.0 \\
\midrule
Mistral-7B & 1K & 92.9 & 92.9 & 92.9 & 93.0 \\
 & 4K & 90.7 & 90.7 & 90.8 & 90.8 \\
\midrule
Falcon3-7B & 1K & 94.4 & 94.5 & 94.5 & 94.5 \\
 & 4K & 91.2 & 91.2 & 91.3 & 91.2 \\
\midrule
Granite-3.1-8B & 1K & 81.0 & 81.0 & 81.0 & 81.0 \\
 & 4K & 78.7 & 78.7 & 78.7 & 78.7 \\
\bottomrule
\end{tabular}
\renewcommand{\arraystretch}{1.0}        
\end{table}

%% file: table4_downstream.tex
\begin{table}[t]
\centering
\scriptsize
\renewcommand{\arraystretch}{1.0}
\caption{Downstream task accuracy (\%). Each cell reports HellaSwag 10-shot / ARC-Challenge 25-shot / WinoGrande 5-shot / average.}
\label{tab:downstream}
\begin{tabular}{l c c c c}
\toprule
\textbf{Model} & \textbf{Baseline} & \textbf{KIVI k2v2} & \textbf{Scalar k3v2} & \textbf{E8 k2v2} \\
\midrule
\multicolumn{5}{c}{\footnotesize Cell format: HellaSwag / ARC-C / WinoGrande / Avg} \\
\midrule
Qwen2.5-14B
& 65.5 / 60.5 / 75.1 / 67.1
& 65.7 / 60.3 / 75.3 / 67.1
& 65.7 / 60.4 / 75.3 / 67.1
& 65.7 / 60.3 / 75.3 / 67.1 \\
Llama-3.1-8B
& 59.8 / 53.8 / 73.6 / 62.4
& 59.7 / 53.6 / 73.4 / 62.2
& 59.7 / 53.7 / 73.4 / 62.3
& 59.7 / 53.7 / 73.4 / 62.3 \\
Mistral-7B
& 65.8 / 57.8 / 74.9 / 66.2
& 65.8 / 57.8 / 74.8 / 66.1
& 65.8 / 57.8 / 74.8 / 66.1
& 65.8 / 57.8 / 74.8 / 66.1 \\
Falcon3-7B
& 60.3 / 55.8 / 70.3 / 62.1
& 60.3 / 55.9 / 70.4 / 62.2
& 60.3 / 55.9 / 70.4 / 62.2
& 60.3 / 56.0 / 70.4 / 62.2 \\
Granite-3.1-8B
& 64.0 / 51.4 / 75.5 / 63.7
& 64.0 / 51.7 / 74.8 / 63.5
& 64.0 / 51.7 / 74.8 / 63.5
& 64.0 / 51.7 / 74.8 / 63.5 \\
\bottomrule
\end{tabular}
\renewcommand{\arraystretch}{1.0}
\end{table}

%% file: table5_ablation.tex
\begin{table}[htbp]
\centering
\caption{Compact ablations (mean across core models). (a) E8 scale calibration. (b) Initial-layer precision sweep.}
\label{tab:ablation}
\small
\begin{minipage}[t]{0.60\linewidth}
\centering
\textbf{(a) E8 scale $\alpha$}\\
\vspace{1pt}
\resizebox{\linewidth}{!}{%
\begin{tabular}{lcccccccc}
\toprule
$\alpha$ & 1 & 4 & 8 & 10 & 12 & 15 & 20 & 30 \\
\midrule
MSE & 0.0577 & 0.0045 & 0.0011 & 0.0008 & \textbf{0.0007} & 0.0008 & 0.0013 & 0.0024 \\
cos & 0.292 & 0.727 & 0.913 & 0.940 & \textbf{0.953} & 0.960 & 0.959 & 0.952 \\
\bottomrule
\end{tabular}%
}
\end{minipage}
\hfill
\begin{minipage}[t]{0.37\linewidth}
\centering
\textbf{(b) Initial-layer bitwidth}\\
\vspace{1pt}
\resizebox{\linewidth}{!}{%
\begin{tabular}{lccccc}
\toprule
Cfg & E8-2b & E8-f4-3b & E8-f8-3b & Sc-3b & Sc-f4-4b \\
\midrule
cos & \textbf{0.954} & 0.952 & 0.950 & 0.920 & 0.927 \\
\bottomrule
\end{tabular}%
}
\end{minipage}
\end{table}